# ROBÓTICA MÓVEL E INTELIGÊNCIA ARTIFICIAL PARA INVESTIGAÇÃO, COMPETIÇÃO E AUTOMATIZAÇÃO DE SISTEMAS INDUSTRIAIS

Jacobs Sodre Pereira, Hiago, hiago.sodre@utec.edu.uy[1]
Moraes Furik, Pablo Ezequiel, pablo.moraes@estudiantes.utec.edu.uy[2]

Grando Bedin, Ricardo, ricardo.bedin@utec.edu.uy[1]
Da Silva Kelbouscas, André, andre.dasilva@utec.edu.uy[2]

[1]Universidad Tecnológica del Uruguay, Rivera, Uruguay
[2]Universidad Tecnológica del Uruguay, Rivera, Uruguay

[1]Universidad Tecnológica del Uruguay, Rivera, Uruguay
[2]Universidad Tecnológica del Uruguay, Rivera, Uruguay

**ABSTRACT**

The implementation of robots to enhance some processes has become popular in recent years due to the accelerated way of production in some factories. Within this context was where robotics has emerged, firstly with stationary robots and more recently mobile robots, namely aerial and terrestrial robots. They can be used for delimited processes within a function, mainly the stationary robots, but also for research in wider areas and even competition. This work summarizes the construction of a model of terrestrial mobile robot that makes the use of artificial intelligence for the purpose of research and competitions, all of that with the basic sensing that can be used in industry.

**RESUMO**

A implementação de robôs para aprimorar alguns processos tornou-se popular nos últimos anos devido à forma acelerada de produção em algumas fábricas, tornou-se necessário substituir a mão de obra humana por processos robóticos. Foi nesse contexto que surgiu a robótica, primeiramente com robôs estacionários e mais recentemente com robôs móveis, nomeados robôs aéreos e terrestres. Eles podem ser usados para processos delimitados dentro de uma função, principalmente os robôs estacionários, mas também para pesquisas em áreas mais amplas e até competitivas. Este trabalho resume a construção de um modelo de robô móvel terrestre que faz uso da inteligência artificial para fins de pesquisa e competições, tudo isso com sensoriamento básico que pode ser utilizado na indústria..

***Palavras-chave:*** *Robôs Móveis Terrestres, Inteligência Artificial, Visão Computacional.*

## 1 - INTRODUÇÃO

O uso de inteligência artificial em adaptações de tecnologia torna perceptível todas as potencialidades transformadoras dessa ferramenta que rompe limites além da robótica. Se analisar os fatores industriais de como ela impacta o aceleramento de produção tornando a competitividade desse ramo cada vez mais alta, máquinas que a uma década atrás não utilizam essa funcionalidade de uma memória de rotina de fabricação, por exemplo, vão em direção a se tornar ultrapassadas.

Aliado a isso, sistemas digitais estão seguindo o mesmo caminho que tecnologias anteriores como carvão à vapor ou eletricidade, robôs estão tomando espaço em nossa sociedade, em distintas faixas etárias, desde crianças e adolescentes; existem modelos de criação nesse contexto até já existem sistemas didáticos para um aprendizado e aperfeiçoamento de habilidades. Para pessoas com uma idade avançada, os projetos tendem a envolver a utilização desses sistemas para cuidados pessoais e domésticos sem a necessidade de uma pessoa ao fazer atividades básicas de suas rotinas diárias, por exemplo.

O fator de existência de um ramo focado em pesquisa e desenvolvimento sobre robôs torna relevante o seu impacto na economia mundial. Desde que a empresa Unimation, por exemplo, instalou o primeiro robô industrial em 1961, na atualidade mais de 700.000 robôs são empregados nas indústrias em todo o mundo. Com respeito a sua estrutura, um robô pode ser definido tradicionalmente como um sistema mecânico, de geometria variada, formada por corpos rígidos, articulados entre si, destinado a sustentar e posicionar/orientar o órgão terminal, que dotado de garra mecânica ou ferramenta especializada, fica em contato direto com o processo. Em algumas aplicações os processos alcançam 100% de robotização, tal como na manipulação de materiais diversos, soldagem por resistência por pontos e pintura na indústria automobilística (Bastos, 2014).

Este projeto apresenta uma aplicação da inteligência artificial ao controle de um tipo de robôs emergentes, os robôs móveis Diferentemente dos tradicionais robôs estacionários, eles conseguem se organizar de forma autônoma e realizar tarefas de movimento no espaço sem estar fixo a um determinado lugar. Esses robôs são muito usados em competições buscando a evolução de novos sistemas, contando com regras de esportes convencionais seguindo modelos já conhecidos. A equipe envolvida nesse projeto cumpre o processo de desenvolver os robôs e colocá-los funcionais para as competições que já existem. A finalidade delimitada a este projeto é os robôs serem controlados por sistemas de inteligência artificial, podendo assim tomar decisões próprias em tarefas previamente programadas, que no contexto desse trabalho é uma partida de futebol.

## 2 - METODOLOGIA

Este projeto envolve a construção de robôs de uma categoria específica de competição com reconhecimento mundial, baseado em modelos existentes no estado da arte. Esses robôs têm o básico do sensoriamento necessário para usá-los em sistemas industriais, tal qual como um lidar, câmeras e outros sistemas de sensoriamento de movimento. A equipe envolvida neste projeto conta com integrantes que trabalham na parte de eletrônica, programação e construção de robôs. O laboratório da equipe se encontra no polo ITR NORTE da Universidade Tecnológica do Uruguai (UTEC), em Rivera, no Uruguai. Neste local é possível a partir de um espaço, ter a infraestrutura necessária para que o projeto seja desenvolvido de maneira conjunta, com equipamentos necessários.

A aplicação final em questão é a Robocup, competição que tem a iniciativa de pesquisa na área de robótica e que busca o desenvolvimento tecnológico por meio de competições em que são utilizados robôs autônomos. Uma dessas competições é o futebol de robôs, que por sua vez é dividido em cinco ligas, uma delas é o SSL para sua sigla em inglês (Small Size League) ou também conhecido como F180, essas que acontecem em etapas nacional, continental e mundial, em diversos países do mundo inteiro. (Rovas, 2015). A equipe participará no próximo ano em uma competição regional simplificada e inspirada na RoboCup, desenvolvida em parceria com a Universidade Federal de Rio Grande (FURG) A competição acontecerá no polo ITR norte da Utec, sendo assim possível provar os robôs já desenvolvidos durante este ano em uma competição demonstrativa ao público. Será possível então provar erros e novas implementações para a competição nacional que a equipe deseja participar em anos seguintes.

Além do objetivo de competição para este projeto, existe o fomento à investigação a ser desenvolvida no laboratório da equipe novas formas de construção destes robôs. As ideias dos estudantes e docentes são integradas como novas possibilidades de adaptação ao projeto, tendo assim, o enriquecimento do conhecimento nesta área avança de forma que torna possível participações e parcerias que rompem fronteiras, com conexões em grandes universidades europeias como a Ostfalia University of Applied Sciences, na Alemanha, que tem experiência de mais de 10 anos nestas categorias.

Para a realização de todos esses objetivos a equipe conta com uma parceria com a embaixada dos Estados Unidos da América de Montevideo, Uruguai, aliado a Universidade Tecnológica do Uruguai (UTEC). O projeto aprovado no início de 2022 conta com o investimento que incentiva o desenvolvimento dos robôs SSL para competição e investigação. Essa parceria inicial possibilitou à equipe investir em equipamentos adequados e todas as peças necessárias para o desenvolvimento dos robôs, chegando a um nível comparado às equipes já existentes.

**2.1 - ROBÓTICA MÓVEL**

Os robôs para este projeto são inspirados em um padrão já existente e delimitado pelas competições. As regras SSL da RoboCup, por exemplo, delimitam que o robô deve caber dentro de um círculo de 180 mm de diâmetro e não deve ter mais de 15 cm de altura. Os robôs jogam futebol com uma bola de golfe laranja em um campo acarpetado verde com de 12 m de comprimento por 9 m de largura na primeira divisão e 9m por 6m na segunda divisão (Robocup, 2020). Isso tudo permite que se mantenha uma linha de construção para todos os competidores e que tenha o desafio de adaptar as necessidades de seu funcionamento para a competição. O modelo de construção deste tipo de robô com uma vista construtiva pode ser visto na Figura 1.

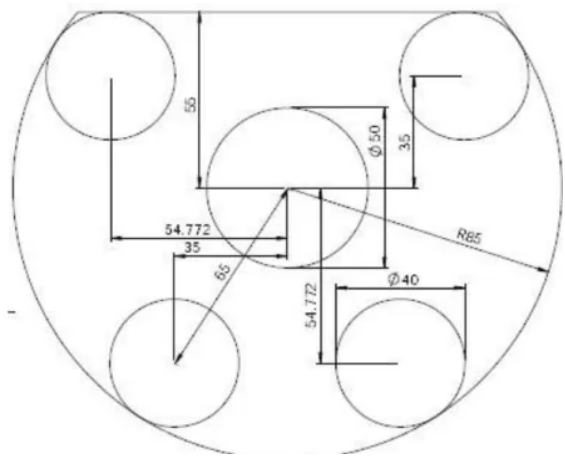

Figura 1. Padrão de construção Robocup 2015. (Rovas, 2015).

Isso tudo permite que se mantenha uma linha de construção para todos os competidores e que tenha o desafio de adaptar as necessidades de seu funcionamento para a competição, aliado com as distintas tecnologias utilizadas a competição chega a inovações todos os anos. Com o surgimento de novas estratégias para a competição cria-se a necessidade de adequar suas regras com limitações aos competidores, isso é, existir uma limitação indireta a criatividade das equipes mantendo uma possibilidade de competição a todos.

Com o avanço do conhecimento nesta área, as equipes se tornam mais capacitadas para desenvolver sistemas mais rápidos, práticos e adaptáveis, isso se vê a cada edição das competições com campeões cada vez mais capacitados. É possível encontrar modelos na comunidade de investigação desta área, porém é necessário adequar se a padrões de valores para o seu projeto, equipes mais desenvolvidas utilizam motores com faixas em 500 € que são mais rápidos e adaptáveis. O maior desafio dentro destas questões, é adaptar o seu melhor modelo às limitações da competição que irá participar com isto, o projeto se torna cada vez mais interessante.

A partir das limitações impostas ao projeto foi pensado em um modelo de construção sobre camadas, onde cada uma comportaria uma parte do controle do robô, componentes eletrônicos, drivers, alimentação, raspberry pi 3, um microcontrolador e tudo que envolve o funcionamento do mesmo. Neste exemplo representado na Figura 2, é perceptível que este modelo de construção é o mais eficaz utilizado para este tipo de projeto, comportando todas as necessidades do robô e tendo divididas as partes de acesso aos controles. Este modelo já foi comprovado como útil por equipes campeãs de categorias internacionais e mundiais, mostrando assim sua eficiência.

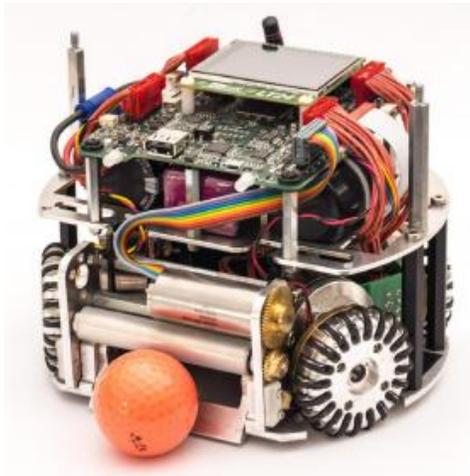

Figura 2. Exemplo de robô categoria SSL. (Robocup, 2020).

Com o intuito de tornar este modelo de robô automatizável, em suas camadas foram inseridos circuitos eletrônicos e controladores que permitem o funcionamento do mesmo. Para este caso, o Raspberry Pi 3 citado, é uma placa multiplataforma ideal, onde é possível utilizá-la como um "mini-computador" próprio para cada robô assim com sua disponibilidade de integração de sistema operacional, neste caso o Linux, torna possível enviar as informações necessárias ao microcontrolador em questão através do sistema operacional de robôs chamado ROS. O microcontrolador utilizado na primeira versão deste robô será a OpenCR, uma placa da empresa Dynamixel que já trabalha com modelos para robôs e drones, disponibilizando interfaces de controles cada vez mais atualizadas.

Tornando os robôs competitivos é necessário agregar dispositivos de controle da bola no momento da partida, como os robôs simulam jogadores em uma partida, é necessário ter os sistemas de defesa e ataque da equipe. Para que estes robôs se tornem jogadores necessitam poder chutar a bola e conduzi-la consigo, os dispositivos de "Kicker" e "Dribler" são necessários nesta questão. O Dribler consiste em rolamentos que permitem manter a bola controlada com o robô enquanto ele avança com a mesma, já o Kicker, permite que o robô chute a bola em uma direção funcionando com a descarga de um circuito capacitivo. Toda essa organização das funções a cada robô será citada mais a frente do projeto na área de controle por inteligência artificial.

A primeira versão realizada deste robô irá funcionar com motores XL430W2050T, da linha Dynamixel da empresa Robotis, sendo da mesma empresa que a placa OpenCR citada anteriormente, tornando possível uma comunicação destes motores com a placa em questão. A utilização desses motores foi feita pelos motivos de adaptação à necessidade, quando se trabalha com robótica e se inicia um projeto, o primeiro objetivo se torna realizar um robô modelo testando todas as possíveis adaptações futuras para as próximas réplicas do mesmo. Neste nosso caso, o motor em questão nos disponibiliza o torque necessário além de uma boa comunicação com softwares e controladores.

O primeiro protótipo construiu-se com a utilização de peças impressas em uma impressora 3D Creality Ender 3 Macrotec, assim foi possível desenvolver as camadas que constituem nosso primeiro robô. Engrenagens e polias para as rodas deste motor foram feitas com peças a corte a laser, em uma máquina especializada, ao total são quatro rodas com seus respectivos motores, alinhados em 45º dos eixos principais deste motor. Nosso sistema de engrenagens do robô foi projetado para aceleração do mesmo, já que os motores dynamixel contam com uma redução embarcada. Decidimos utilizar esses motores pelo alto torque e eficiência deles, contando com uma segurança de funcionamento já provada em outros projetos da empresa Robotis. O primeiro protótipo realizado ainda esse ano se encontra nas figuras 3 e 4.

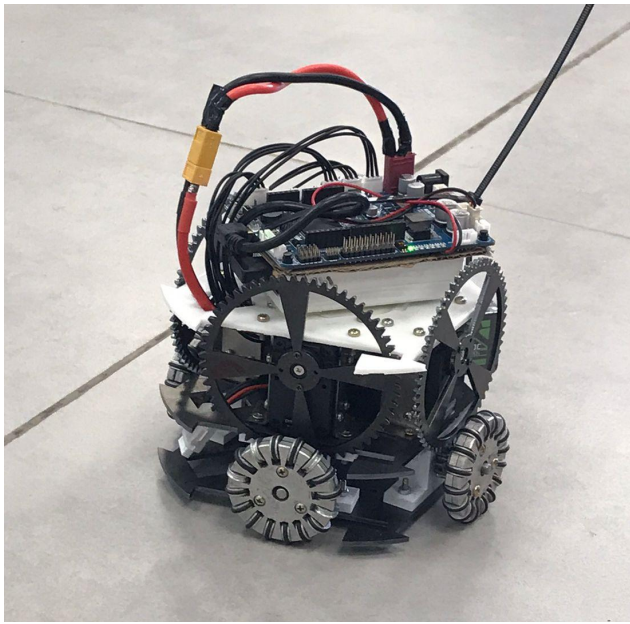 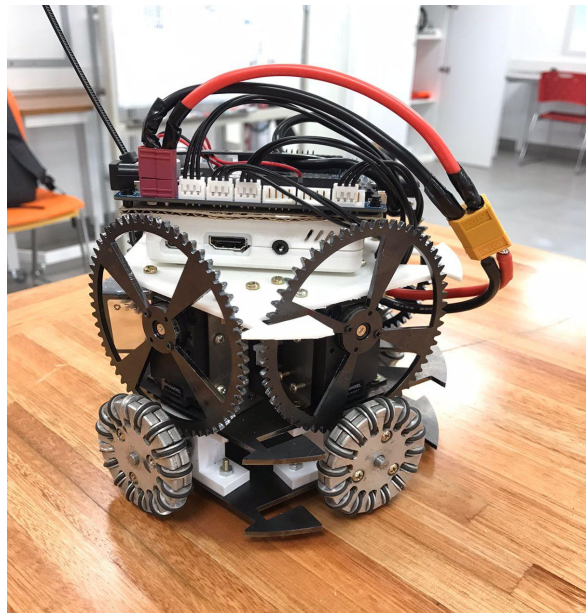

Figura 3. Primeiro protótipo de robô SSL.     Figura 4. Primeiro protótipo de robô SSL.

## 2.2 - INTELIGÊNCIA ARTIFICIAL

A utilização deste tipo de ferramenta para controles e situações de organização de sistemas, se torna interessante com a possibilidade de autonomia dos mesmos. Os sistemas de inteligência artificial possibilitam que os robôs se adaptem a certas situações de uma maneira mais autônoma, é possível assim, prever situações e sugerir opções para uma escolha quase instantânea dos robôs. Para a categoria em questão que atribui uma competitividade grande, é necessário uma tomada de decisão precisa e rápida em certos momentos, esse sistema facilita esse processo e traz consigo a inovação do funcionamento autônomo dessas máquinas.

Uma forma interessante para esta aplicação é com sistema especialista, que é baseado no conhecimento e foi especialmente projetado para emular a especialização humana de algum domínio específico. Este sistema foi construído por uma base de conhecimento formada de fatos, regras e heurísticas sobre o domínio, tal como um especialista humano faria, e deve ser capaz de oferecer sugestões e conselhos aos usuários e, também, adquirir novos conhecimentos e heurísticas com essa interação (BARONE,2003).

Todo funcionamento destes comandos de inteligência artificial com programação foi feita na linguagem PYTHON, desenvolvida com intuito de facilidade, esta linguagem inclui diversas estruturas de alto nível (listas, dicionários, complexos e outras) e uma vasta seleção de módulos prontos para uso além de frameworks de terceiros que podem ser adicionados. (Borges, 2014). Aliado a essa linguagem foi possível utilizar o ROS citado anteriormente, que é uma ferramenta de adaptação e envio de programas de diferentes linguagens de programação através de sistemas conjuntos. ROS possibilita que enviemos aos robôs diversos tipos de programas executando-os diretamente do terminal de comando do Linux que está inserido dentro de nosso Raspberry Pi, o cérebro do nosso robô. Na Figura 5, é possível ver como se forma a utilização deste sistema de controles dos quatro robôs dentro do espaço destinado ao esporte.

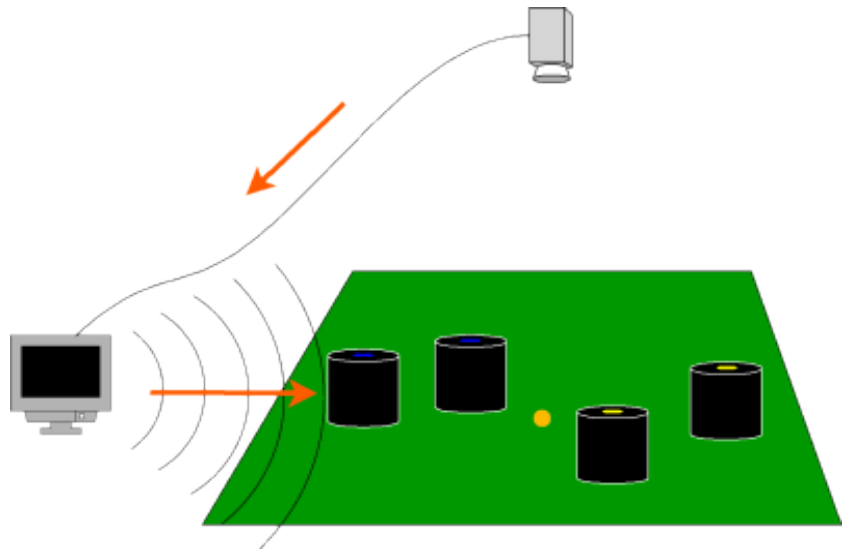

Figura 5. Ilustração do sistema de controle com Inteligência Artificial. (Robocup, 2020).

Acessando de maneira remota esse "cérebro" é possível enviar os programas e testar funcionalidades dos motores de forma rápida e usual. Cada robô se localiza dentro do local da competição por cores que são inseridas em cima deles, como visto na figura 6, são as mesmas cores para eles porém isso não interfere na localização pois o que o software capta com a câmera, são suas posições dentro da quadra. A bola utilizada por exemplo, é de cor laranja, isso como definição da competição para o sistema de localização poder localizá-la e calcular sua posição dentro da quadra. Esse sistema de posição envolve ângulos e distâncias, e por isso, envolve cálculos físicos e matemáticos que são inseridos dentro das linhas de programação, disponibilizando ao sistema ferramentas de cálculo para ele obter dados em questão de segundos.

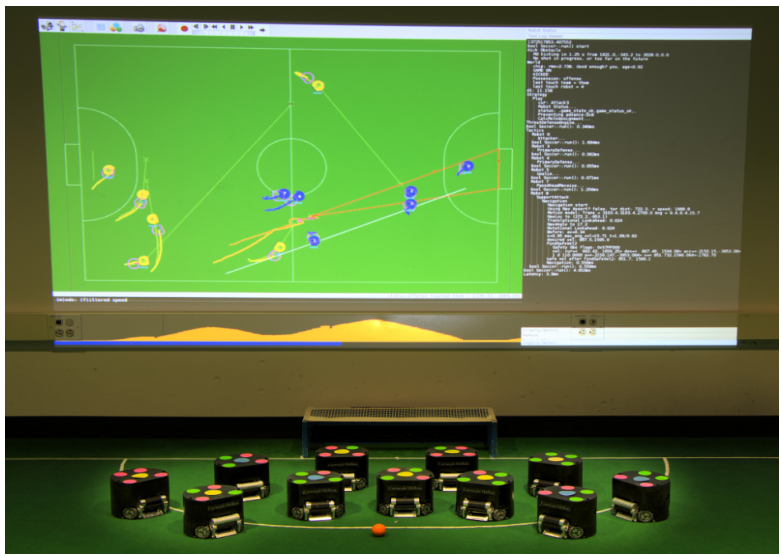

Figura 6. Robôs equipe CMDragons. (CMDragons, 2010).

Cada robô assim, deverá com os dados disponibilizados a ele calcular posição da bola e direção do gol, escolhendo a melhor alternativa para chegar ao objetivo, ele deve recalcular cada movimento em caso de mudança. Por esses motivos, este modelo de competição se torna complexo a nível que podem existir equipes com os robôs mais robustos que aguentem impactos e grandes velocidades, porém se seu sistema de inteligência artificial entre os robôs não estiver bem desenvolvido poderá perder para uma equipe que realiza táticas melhores. Por isso, como no esporte que está aplicado nessa categoria, tática e habilidade necessitam estar em equilíbrio para um melhor resultado, dessa forma se cria uma organização com uma hierarquia para as funções de cada robô.

A maneira em que se seleciona a organização se dá por opções, um robô que funciona de maneira mais ágil, por exemplo, se coloca à disposição do gol podendo buscar os chutes com mais precisão. Já para a parte do ataque se calcula qual robô está mais próximo da bola ou da posição futura dela, enviando assim o mais adequado a esta coordenada. Todas essas previsões são feitas para que a equipe tenha um melhor desempenho, utilizando a integração da inteligência artificial com o sistema para robôs (ROS) facilita o processo de idéias se tornarem possíveis em um projeto como este.

## 3 - CONCLUSÃO

Sobre os olhos de quem realizou este projeto, é possível perceber o quão impressionante foi o avanço em questão de meses trabalhando no mesmo. O empenho de toda a equipe em torno das possíveis resoluções de problemas que se encontravam durante o processo foi primordial para que em questão de pouco tempo, se obtivesse um resultado satisfatório. Não se pode também descartar o conhecimento e interesse que vem de pessoas que alavancaram esta ideia, foi necessário ter instrutores com interesse real neste projeto, motivando a conclusão do mesmo.

Foi possível a partir dos conhecimentos adquiridos com o desenvolvimento deste projeto, aprender e se interessar mais por esta área de investigação, a robótica é um âmbito do conhecimento necessário para o desenvolvimento humano. A maneira com que se integra o uso destas tecnologias a uma aplicação de nível avançado como esta, demonstra o quão interessante e revolucionária esta área é, todo novo descobrimento em questão de visão, controle, mapeamento e computação será implementado a este tipo de categoria de competição, sem descartar o fato do surgimento de novas categorias de robôs no futuro. Se espera assim, que o interesse de investimentos neste setor de tecnologia siga crescendo como se viu nos últimos anos, pois quanto mais as grandes empresas vêm oportunidades neste segmento, cresce o fortalecimento da comunidade acadêmica, podendo seguir desenvolvendo tecnologia e capacitando grandes mentes em projetos brilhantes.

A equipe espera continuar com os trabalhos neste projeto, ao final deste ano ocorrerá uma competição demonstrativa no polo ITR norte da Utec, em conjunto com universidades parceiras, uma partida demonstrativa da categoria SSL será realizada aberta ao público. Se espera que até lá já tenha ao menos 5 robôs construídos e funcionando para que ocorra uma partida bem sucedida. Após isso a equipe tem idéias futuras de adaptação deste projeto a componentes melhores, alguns motores brushless que outras equipes já utilizam e que comportam um torque melhor já foram requisitados, esses motores serão adaptados a esse projeto tornando os robôs cada vez mais competitivos para competições futuras.

## 4 - APOIOS EXTERNOS

Este projeto só foi possível ser realizado com o apoio externo de investimentos acadêmicos. O primeiro apoio foi fruto de uma parceria realizada entre a Embaixada dos Estados Unidos da América em Montevideo, Uruguai, e a Universidade Tecnológica do Uruguai (UTEC), em Rivera, Uruguai. Por iniciativa do docente Ricardo Bedin Grando que ministra aulas na Utec, a parceria de apoio ao projeto de construção dos robôs SSL recebeu um um investimento de USD 20.000 sendo esse apoio direcionado a compra de materiais e equipamentos necessários para o desenvolvimento do projeto.

Além do apoio de investimentos, a Utec tem grandes projetos em parceria com a Universidade Federal do Rio Grande (FURG), alguns cursos ministrados na Utec são em conjunto com ela e esse projeto a Furg apoia com seu conhecimento e equipamentos para a construção desde o início, que são enviados diretamente de Rio Grande no Brasil, para o fortalecimento desta parceria. Como já citado anteriormente, a equipe que desenvolve esse projeto se encontra com um laboratório da Utec em Rivera, então só foi possível a realização desta ideia com o apoio e infraestrutura da Utec, tornando viável o investimento de todo tempo e conhecimento.